\documentclass[a4paper,conference]{IEEEtran}
\IEEEoverridecommandlockouts
\usepackage{cite}
\usepackage{amsmath,amssymb,amsfonts}
\usepackage{algorithmic}
\usepackage{graphicx}
\usepackage{textcomp}
\usepackage{xcolor}
\usepackage{multirow}
\usepackage{subcaption}
\def\BibTeX{{\rm B\kern-.05em{\sc i\kern-.025em b}\kern-.08em
    T\kern-.1667em\lower.7ex\hbox{E}\kern-.125emX}}
\begin{document}

\newcommand{\WSPolypauthorcell}[3]{%
  \parbox[t]{0.235\textwidth}{\centering
    \fontsize{11}{12.5}\selectfont
    \strut #1\\[-0.1em]
    \fontsize{9.2}{10.5}\selectfont
    \strut\mbox{\textit{#2}}\\[-0.05em]
    \fontsize{9.2}{10.5}\selectfont
    \strut\mbox{#3}\par
  }%
}

\makeatletter
\newcommand{\twocolumnfootnotefullwidth}[1]{%
  \begingroup
  \renewcommand{\thefootnote}{}%
  \footnotetext{%
    \noindent\hspace*{-1em}\rule{0.3\linewidth}{0.4pt}\par
    \vspace{0.3em}
    \noindent\footnotesize #1
  }
  \endgroup
}
\makeatother

\title{WSPolypNet: Weakly Supervised Polyp Localization in Colonoscopy Videos}

\author{

\makebox[\textwidth][c]{%
\WSPolypauthorcell
{Giseong Hwang\textsuperscript{*}}
{Soonchunhyang University}
{hwanggiseong@sch.ac.kr}%
\hfill
\WSPolypauthorcell
{Minjae Jo\textsuperscript{*}}
{Seoul National University}
{whalswo0503@snu.ac.kr}%
\hfill
\WSPolypauthorcell
{Yeonghyeon Park}
{Kyungpook National University}
{yeonghyeon05@knu.ac.kr}%
\hfill
\WSPolypauthorcell
{Kyeonghun Kim}
{OUTTA}
{kyeonghun.kim@outta.ai}%
}%

\\[1.2ex]

\makebox[\textwidth][c]{%
\WSPolypauthorcell
{Seoyeon Han}
{Seoul National University}
{ss\_stella@snu.ac.kr}%
\hfill
\WSPolypauthorcell
{Donghoon Han}
{Seoul National University}
{leukinii@snu.ac.kr}%
\hfill
\WSPolypauthorcell
{Haneul Kim}
{Chung-Ang University}
{kimhn126@cau.ac.kr}%
\hfill
\WSPolypauthorcell
{Yului Jeong}
{Seoul National University}
{yule27@snu.ac.kr}%
}%

\\[1.2ex]

\makebox[\textwidth][c]{%
\WSPolypauthorcell
{Insung Hwang}
{Seoul National University}
{insung0608@snu.ac.kr}%
\hfill
\WSPolypauthorcell
{Pa Hong}
{Samsung Changwon Hospital}
{papa.hong@samsung.com}%
\hfill
\WSPolypauthorcell
{Ken Ying-Kai Liao}
{NVIDIA AI Technology Center}
{kenyingkail@nvidia.com}%
\hfill
\WSPolypauthorcell
{Nam-Joon Kim\textsuperscript{\dag}}
{Seoul National University}
{knj01@snu.ac.kr}%
}%

}

\maketitle

\twocolumnfootnotefullwidth{%
\textsuperscript{*} Giseong Hwang and Minjae Jo contributed equally to this work.\\
\textsuperscript{\dag} Corresponding author.
}

\begin{abstract}

Because dense frame-level annotation of colonoscopy videos is costly, we propose WSPolypNet, a weakly supervised framework for polyp localization using only video-level labels. WSPolypNet employs a 3D convolutional neural network trained with video-level labels to generate class activation maps (CAMs) that identify candidate polyp regions without frame-level spatial annotations. The CAM-derived localization cues are enhanced using a multi-view strategy and provided to MedSAM2 as point prompts. These prompts propagate segmentation masks across the video, refining the coarse localization according to polyp boundaries. WSPolypNet achieved CorLoc scores of 47.80\%, 43.68\%, and 35.01\% at IoU thresholds of 0.3, 0.5, and 0.7, respectively, compared with 36.87\%, 33.72\%, and 27.94\% for the single-view setting. For small polyps, multi-view localization improved CorLoc@0.5 from 16.01\% to 30.97\%. The framework also achieved a recall of 94.51\%. These results demonstrate the potential of weakly supervised spatiotemporal learning to reduce spatial annotation requirements in colonoscopy videos.

\end{abstract}

\begin{IEEEkeywords}
polyp localization, weakly supervised learning, video-level supervision, class activation map, MedSAM2
\end{IEEEkeywords}

\section{Introduction}

Colorectal cancer (CRC) is one of the most common and deadly cancers worldwide, accounting for approximately 2 million new cases and nearly 1 million deaths annually. The global burden of CRC is expected to continue increasing, with the number of new cases projected to exceed 3.2 million per year by 2050 \cite{bray2024global}. Because a substantial proportion of CRC cases arise from adenomatous polyps, detecting and removing these precancerous lesions before malignant progression is critically important. Winawer et al. \cite{winawer1993prevention} reported that early detection and removal of adenomatous polyps could reduce the expected incidence of CRC by up to 90\%. Furthermore, the removal of adenomas has been associated with a 53\% reduction in CRC-related mortality compared with that expected in the general population \cite{zauber2012colonoscopic}.

Colonoscopy is widely used as a primary screening and diagnostic procedure for detecting and removing precancerous polyps \cite{corley2014adenoma, bray2024global}. Nevertheless, conventional colonoscopy still has several limitations. Previous studies have reported that as many as 20\%--26\% of polyps may be missed during routine colonoscopic examinations \cite{leufkens2012factors}. In addition, the complex structure of the intestinal wall, the limited field of view, and visual disturbances in endoscopic images can make reliable polyp detection challenging. Consequently, computer-aided detection (CADe) systems have been extensively investigated to assist clinicians and improve the consistency of polyp detection.

Despite these advances, many existing polyp detection methods process frames extracted from endoscopic videos as independent static images. These image-based approaches do not explicitly model temporal correlations between consecutive frames and may therefore produce temporally inconsistent predictions in real-world colonoscopy videos \cite{ji2021progressively}. In particular, complex intestinal structures and visual patterns resembling polyps can make it difficult to distinguish polyps from the surrounding tissue using a single frame. Therefore, reliable and consistent polyp localization in colonoscopy videos requires both spatial information and temporal modeling \cite{belharbi2023tcam}.

Meanwhile, conventional supervised polyp detection and segmentation methods that aim to achieve accurate spatial localization generally rely on detailed spatial annotations, such as bounding boxes or pixel-level masks. Providing such annotations for every frame of an endoscopic video requires substantial time and effort, making it difficult to scale to large video datasets. Moreover, pixel-level annotation can be affected by inter-observer variability because of the complex structure of the intestinal wall and ambiguous polyp boundaries. To reduce this annotation burden, weakly supervised learning approaches have received increasing attention. In particular, Weakly Supervised Video Object Localization (WSVOL) aims to localize objects in videos using coarse supervision, such as video-level labels, rather than dense frame-level spatial annotations \cite{kumar2025explainable, belharbi2023tcam}.

Previous studies have explored the use of video-level supervision to classify the presence of target objects or estimate their temporal locations within videos \cite{belharbi2023tcam, liao2025disentangling}. In parallel, weakly supervised object localization methods have investigated spatial object localization using only image-level supervision \cite{zhou2016learning}. However, spatially localizing polyps in colonoscopy videos using only video-level supervision remains relatively underexplored. Moreover, polyp appearance and viewpoint continuously change throughout endoscopic videos, while complex intestinal structures and various visual disturbances may cause weakly supervised localization models to activate irrelevant regions \cite{choe2019attention}. Therefore, a method is needed that can exploit spatiotemporal information to localize polyps reliably without dense spatial annotations.

To address these challenges, we propose a weakly supervised framework that simultaneously determines the presence of polyps at the video level and estimates their spatial locations in individual frames using only video-level labels. The proposed framework employs a 3D convolutional neural network (3D CNN) to learn spatiotemporal representations from consecutive endoscopic frames and generates class activation maps (CAMs) \cite{zhou2016learning, selvaraju2017grad} to identify candidate polyp regions without additional spatial annotations. The CAM-derived localization cues are subsequently provided to MedSAM2 \cite{ma2025medsam2} as prompts to refine the coarse regions to better align with polyp boundaries. Thus, the proposed framework is designed to estimate both polyp presence and spatial location using only video-level supervision, without requiring frame-level bounding-box or pixel-level annotations.

The main contributions of this study are summarized as follows:

\begin{itemize}

\item We propose a weakly supervised polyp localization framework that simultaneously determines polyp presence at the video level and performs spatial localization in individual frames using only video-level labels.

\item We integrate a 3D CNN with CAMs to exploit spatiotemporal information from consecutive endoscopic frames and estimate candidate polyp regions without frame-level spatial annotations.

\item We construct a localization refinement pipeline that uses CAM-derived localization information as prompts for MedSAM2 to refine coarse localization results according to polyp boundaries.

\end{itemize}

\section{RELATED WORK}

\subsection{AI-based Polyp Detection and Localization}

Deep learning has been widely applied to polyp detection and segmentation in endoscopic images. Representative approaches include real-time CNN-based detection, multi-scale feature fusion and attention mechanisms, and Transformer-based architectures \cite{urban2018deep, xie2023gastric, wang2023vision}. 

However, most existing methods process individual frames independently and often rely on strong spatial supervision, such as bounding boxes or pixel-level masks. Such frame-wise approaches do not fully exploit temporal information across consecutive frames and require substantial annotation effort for long endoscopic videos.

\subsection{Weakly Supervised Object and Video Localization}

Weakly Supervised Object Localization (WSOL) aims to estimate object locations using image-level labels without dense spatial annotations. Representative methods include CAM \cite{zhou2016learning}, Grad-CAM \cite{selvaraju2017grad}, approaches that discover complementary object regions \cite{choe2019attention, zhang2018adversarial}, and Transformer-based localization methods \cite{gao2021ts, gupta2022vitol}.

Weakly supervised localization has also been extended to video using video-level supervision and temporal information, including temporal CAM-based methods, weakly supervised surgical video localization, and Transformer-based video object localization \cite{belharbi2023tcam, liao2025disentangling, murtaza2024leveraging}.

However, CAM-based localization often focuses on only the most discriminative object regions, resulting in incomplete or spatially coarse localization \cite{choe2019attention, zhang2018adversarial}. This limitation is particularly problematic for small polyps, while spatial localization of polyps in colonoscopy videos using only video-level supervision remains relatively underexplored.

\subsection{Promptable Medical Image and Video Segmentation}

Promptable segmentation models such as SAM \cite{kirillov2023segment} generate dense masks from sparse spatial prompts. SAM 2 \cite{ravi2024sam2} extends this capability to video, while MedSAM2 \cite{ma2025medsam2} adapts it to medical image and video analysis.

Because these models require an initial spatial prompt, they are complementary to weakly supervised localization. In our framework, CAM-derived candidate locations are used as point prompts for MedSAM2, which refines the coarse localization into segmentation masks without requiring frame-level spatial annotations for the target training videos.

\section{PROPOSED METHOD}

\subsection{Overall Framework}

The proposed framework performs weakly supervised polyp localization in colonoscopy videos using video-level labels. Given a preprocessed video clip, a 3D convolutional neural network (3D CNN) first extracts spatiotemporal representations and predicts the presence of a polyp at the video level. Class activation maps (CAMs) are then derived from the trained classifier to obtain coarse spatial localization cues without requiring frame-level bounding-box or pixel-level mask annotations.

\begin{figure*}[t]
    \centering
    \includegraphics[width=\textwidth]{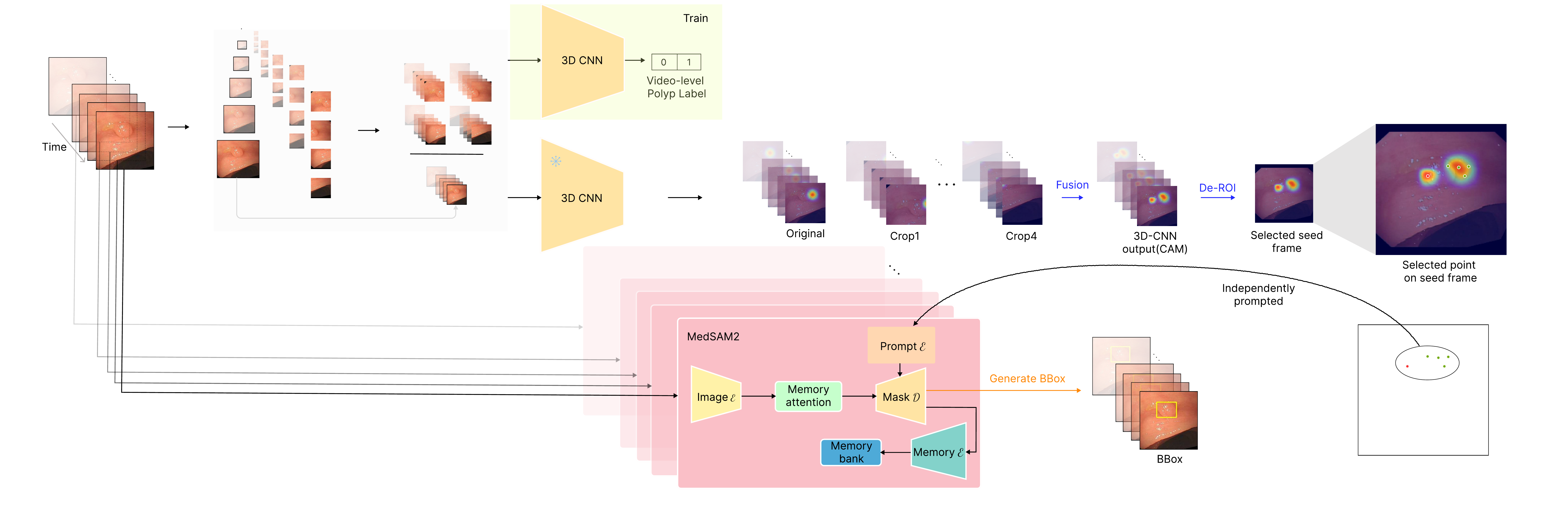}
    \caption{Overall framework of the proposed method. Given an input video clip, a 3D CNN predicts video-level polyp presence and generates class activation maps (CAMs), from which candidate points are derived. Each point is independently provided to MedSAM2 as a prompt, which propagates the resulting mask across the clip to produce the final localization.}
    \label{fig:overall_framework}
\end{figure*}

The CAM-derived localization cues are further enhanced using a multi-view strategy and converted into candidate point prompts for MedSAM2. MedSAM2 propagates the resulting segmentation masks bidirectionally across the video sequence, after which the candidate tracks are evaluated to determine the final localization result. The overall framework is illustrated in Fig.~\ref{fig:overall_framework}.

\subsection{ROI Preprocessing}
\label{sec:roi}

The positive and negative video sources exhibit substantial differences in image resolution and peripheral visual appearance, including black borders and interface overlays. Such source-specific cues may introduce visual bias unrelated to polyp appearance. To mitigate this issue, we apply field-of-view (FOV)-based region-of-interest (ROI) preprocessing to all input frames.

For each video, the effective endoscopic FOV is estimated using temporal brightness voting over frames uniformly sampled across the temporal dimension. The detected FOV is represented by its convex hull, after which the bounding region enclosing the hull is cropped and pixels outside the hull are masked to zero. When substantial variations in FOV shape are observed across temporally separated sampled frames, the union of the detected regions is used to avoid excluding valid endoscopic areas.

Because fixed background patterns and ROI boundaries may themselves serve as source-specific cues, additional spatial randomization is applied during training. The extracted ROI is randomly resized to 90--100\% of the training canvas while preserving its aspect ratio and is placed at a randomly selected valid position. This augmentation reduces the likelihood that the classifier relies on consistent padding patterns or ROI locations rather than polyp-related visual features. During evaluation, random scaling is disabled and the ROI is deterministically placed at the center of the canvas.

\begin{figure}[t]
    \centering
    \includegraphics[width=\columnwidth]{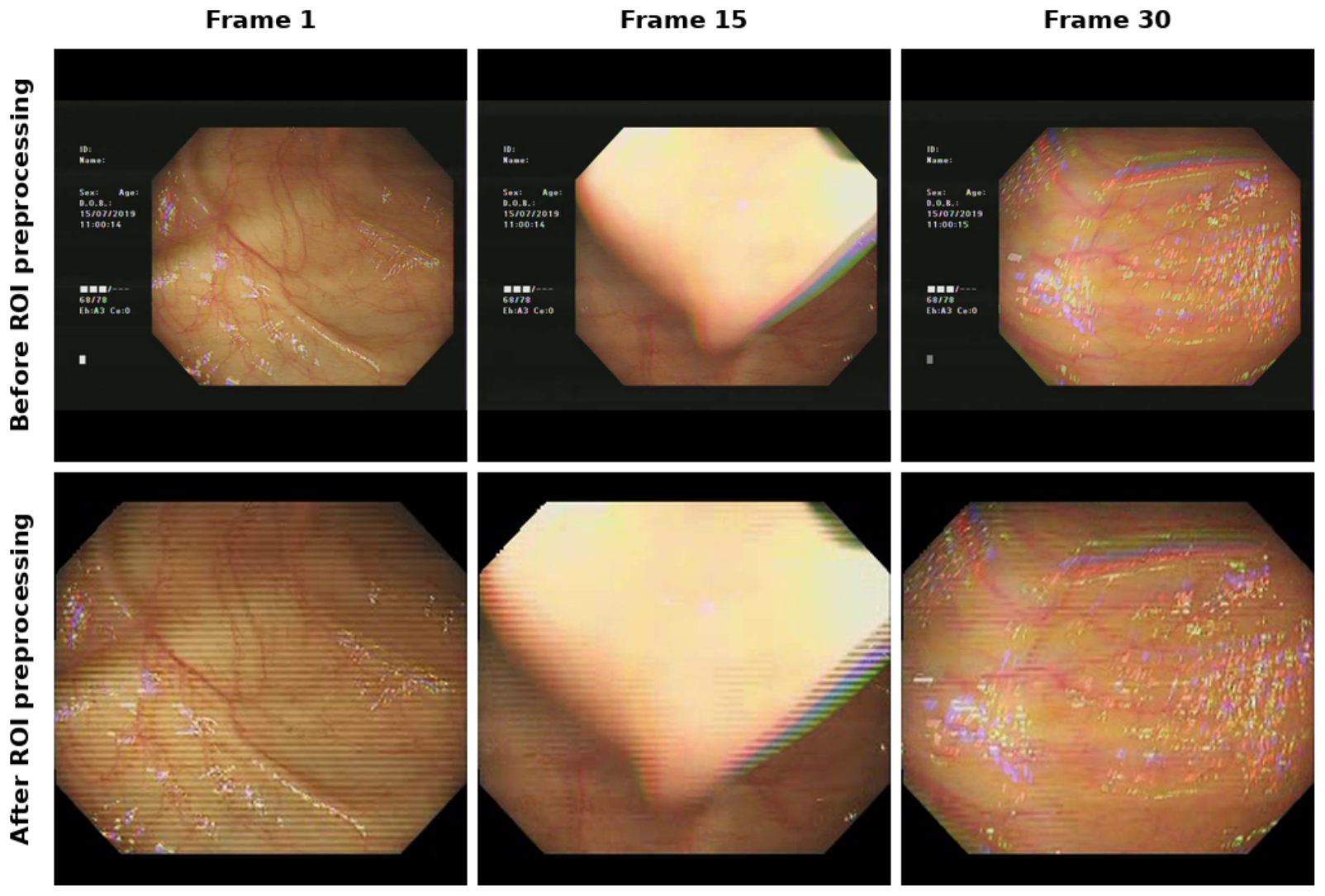}
    \caption{Effect of FOV-based ROI preprocessing on a representative clip. Top: original frames. Bottom: ROI-preprocessed frames.}
    \label{fig:roi}
\end{figure}

This preprocessing removes irrelevant peripheral regions while preserving the effective endoscopic field of view and visible mucosal area as much as possible. Fig.~\ref{fig:roi} shows a representative clip before and after ROI preprocessing.

\subsection{Spatiotemporal Feature Learning and CAM Generation}

Let an input video clip be denoted by
$\mathbf{X} \in \mathbb{R}^{C \times T \times H \times W}$.
The clip is fed into a 3D CNN to extract spatiotemporal feature representations from consecutive endoscopic frames. The final convolutional feature tensor is denoted by
$\mathbf{F} \in \mathbb{R}^{K \times T' \times H' \times W'}$,
where $K$ is the number of feature channels and $T'$, $H'$, and $W'$ denote the temporal and spatial dimensions of the feature tensor, respectively.

For video-level polyp classification, global average pooling (GAP) is applied to $\mathbf{F}$, and the resulting feature vector is passed through a binary classification layer that produces a single logit. The classifier is trained using a video-level binary label $y \in \{0,1\}$ and binary cross-entropy with logits loss (BCEWithLogitsLoss). During inference, the output logit is converted to a polyp-presence probability using a sigmoid function.

To obtain spatial localization cues from the trained classifier, a class activation map (CAM) is computed from the final convolutional feature tensor. Let $w_k$ denote the classification weight associated with the $k$-th feature channel. The CAM is defined as

\begin{equation}
M(t,u,v)
=
\sum_{k=1}^{K}
w_k F_k(t,u,v),
\end{equation}

where $F_k(t,u,v)$ denotes the activation of the $k$-th feature channel at temporal index $t$ and spatial location $(u,v)$. The resulting class-specific response map indicates the contribution of each spatiotemporal location to the polyp-present prediction and serves as the initial localization cue for the subsequent refinement stage.

\subsection{MedSAM2-based Localization Refinement}

To improve localization, particularly for small polyps, the CAM evidence of each frame is aggregated over five views: the full 224×224 model input and four overlapping corner crops. Each corner crop is a 144×144 window anchored at one corner of the input (covering 64\% of each side, so that adjacent crops overlap in a 64-px-wide central band). Every crop is bilinearly resized back to 224×224 — a $\approx$1.56× magnification — and passed through the same frozen classifier, so that polyps near the field-of-view boundary, which the full view under-resolves on the coarse feature grid, are enlarged before the CAM is computed. A centered crop was also examined but gave no improvement and is not used. For each view $q$, let $A_t^{(q)}(u,v)$ denote the raw CAM response at frame $t$. The corresponding localization score map is computed as

\begin{equation}
L_t^{(q)}(u,v)
=
S_t^{(q)}(u,v)
\cdot
\sigma\left(A_t^{(q)}(u,v)\right),
\end{equation}

where $\sigma(\cdot)$ denotes the sigmoid function and
$S_t^{(q)}(u,v)$ denotes the spatial weight obtained by applying softmax over the spatial locations of the raw CAM response. This formulation combines the relative spatial importance of each location with its activation strength.

The localization maps from the five views are mapped back to the coordinate system of the original frame and fused using a pixel-wise maximum operation:

\begin{equation}
L_t^{\mathrm{fused}}(u,v)
=
\max_{q \in \{1,\ldots,5\}}
L_t^{(q)}(u,v).
\end{equation}

For each frame, the peak response of the fused localization map is defined as

\begin{equation}
s_t
=
\max_{u,v}
L_t^{\mathrm{fused}}(u,v),
\end{equation}

and the seed frame is selected as

\begin{equation}
t^{*}
=
\arg\max_t s_t.
\end{equation}

The fused localization map of the seed frame $t^{*}$ is min--max normalized to $[0,1]$ and then Gaussian smoothed with $\sigma = 4$~px in the $224\times224$ input space. Non-maximum suppression is applied greedily: the current maximum is taken as a candidate point, all locations within a radius of $16$~px are suppressed, and the step repeats until five candidate points are obtained, ordered by decreasing response.

Each candidate point is independently provided to MedSAM2 as a positive point prompt. MedSAM2 generates an initial segmentation mask at the seed frame and propagates the mask bidirectionally to the preceding and subsequent frames, producing one candidate segmentation track for each point.

The candidate tracks are evaluated using the CAM-derived fused localization responses. For each frame, the mean fused localization response is computed within the bounding region enclosing the predicted mask, and the track score is obtained by averaging these frame-level responses over the video sequence. The first-ranked candidate is selected by default, while another candidate replaces it only when its track score exceeds that of the first-ranked candidate by more than 20\%.

For a given frame, the propagated mask is replaced by an independent frame-wise prediction when either (i)~the propagated mask is empty, or (ii)~its confidence falls below a threshold $\tau_{\mathrm{conf}} = 0.755$, where the per-frame confidence is the mean sigmoid probability over the foreground pixels of the mask. Condition~(ii) covers tracking drift, in which the seed frame is localized correctly but the target is lost on other frames of the clip. The frame-wise fallback prediction is obtained by prompting MedSAM2 with a single positive point at that frame's own fused-CAM peak, without using temporal information. Through this refinement process, the coarse CAM-based localization cues are converted into segmentation masks with more precise spatial boundaries.

\section{EXPERIMENTS}

\subsection{Dataset}

For video-level binary classification, we constructed a training dataset consisting of positive and negative video clips. Positive samples were generated from 100 frame sequences extracted from the LDPolypVideo dataset. The frames in each sequence were divided into non-overlapping clips of up to 30 consecutive frames while preserving their original temporal order. This procedure resulted in a total of 871 positive video clips.

Negative samples were generated from 60 original endoscopic videos without polyps. Each negative clip was constructed from 30 temporally consecutive, non-overlapping frames extracted from a single source video. Frames originating from different source videos were not combined within the same clip. This procedure yielded a total of 615 negative video clips.

The final training dataset consisted of 871 positive clips and 615 negative clips, yielding a total of 1,486 video clips. The output frame rate of all generated clips was set to 1 FPS. In addition, to reduce source-specific visual bias arising from differences in image resolution and peripheral appearance between the positive and negative data sources, the ROI preprocessing described in Section~\ref{sec:roi} was applied to all input frames.

\subsection{Implementation Details}

All 3D CNN backbones were initialized with weights pretrained on Kinetics-400 and trained under the same optimization settings. The models were optimized using AdamW with an initial learning rate of $1\times10^{-4}$ and a weight decay of 0.01. The backbone was frozen for the first five epochs and subsequently unfrozen for end-to-end fine-tuning. Training was performed for a total of 50 epochs using a cosine learning-rate schedule.

\begin{figure*}[!t]
    \centering
    \includegraphics[width=\textwidth]{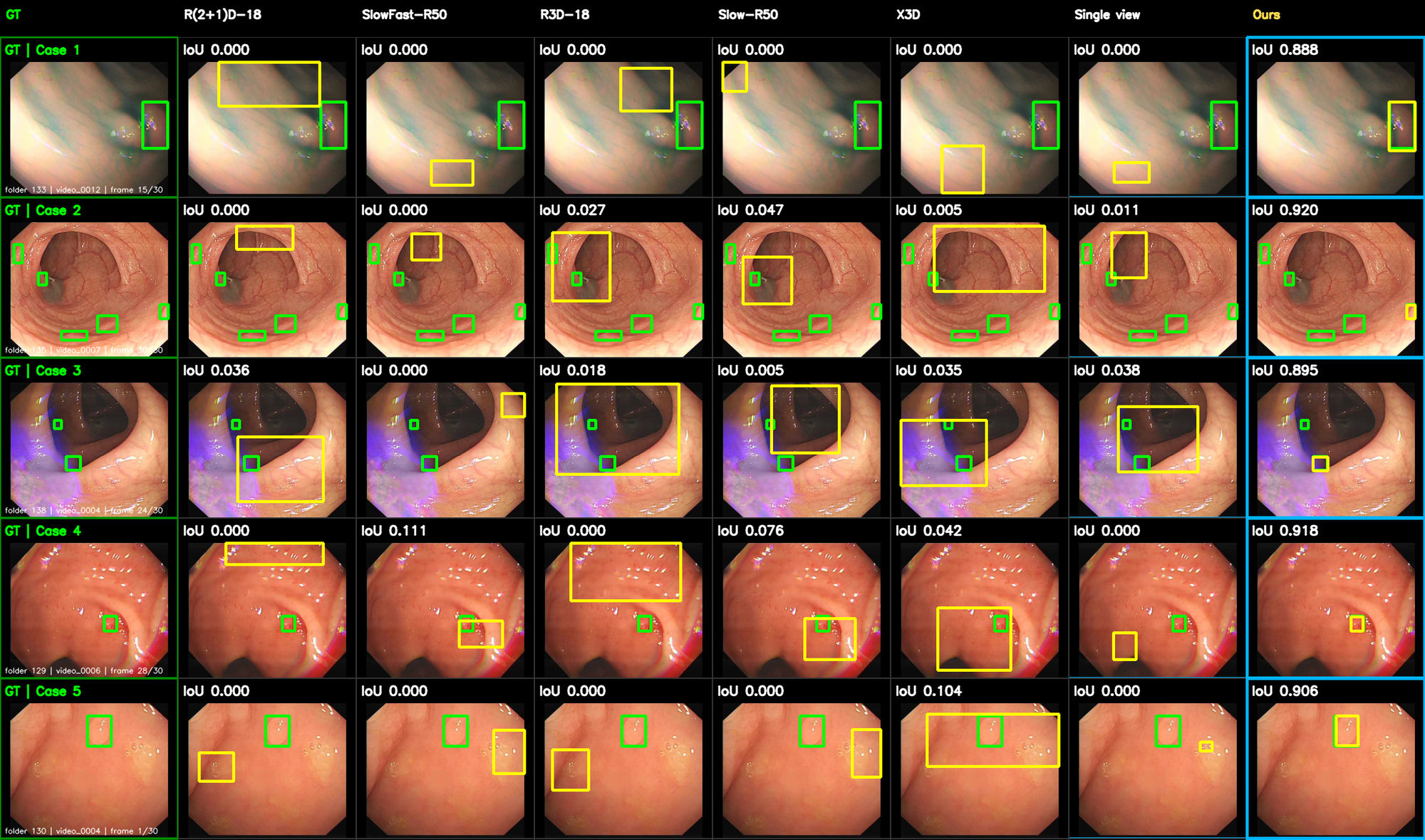}
    \caption{Qualitative localization comparison on representative test frames. Green and yellow boxes denote ground truth and predictions, respectively. Columns compare five 3D CNN backbones, the single-view setting, and the full WSPolypNet framework.}
    \label{fig:qual}
\end{figure*}

\subsection{Evaluation Metrics}

Localization performance was evaluated using Correct Localization (CorLoc). A localization result was considered correct when the intersection over union (IoU) between the predicted and ground-truth bounding boxes exceeded a predefined threshold. CorLoc was evaluated at IoU thresholds of 0.3, 0.5, and 0.7. Recall was additionally reported to evaluate the ability of each method to identify positive samples.

For the analysis of the multi-view strategy, polyps were divided into small and large groups based on the ratio of the ground-truth bounding-box area to the total frame area. Polyps with an area ratio of 0.05 or less were categorized as small, whereas those with an area ratio greater than 0.05 were categorized as large.

\subsection{Experimental Results}

To compare the localization capability of different 3D CNN backbones, five architectures were evaluated under the same training and evaluation settings. Table~\ref{tab:backbone_comparison} summarizes their Recall and CorLoc performance. Representative qualitative results are shown in Fig.~\ref{fig:qual}.

\begin{table}[!t]
\centering
\caption{Performance comparison of different 3D CNN backbones.}
\label{tab:backbone_comparison}
\begin{tabular}{lcccc}
\hline
Backbone & Recall & CorLoc@0.3 & CorLoc@0.5 & CorLoc@0.7 \\
\hline
Slow R50     & 80.22 & 18.94 & 7.87 & 1.05 \\
X3D          & 74.18 & \textbf{21.03} & \textbf{9.32} & \textbf{1.63} \\
SlowFast R50 & 69.23 & 13.57 & 3.63 & 0.59 \\
R3D-18       & 91.21 & 16.75 & 6.53 & 0.82 \\
R(2+1)D-18   & \textbf{94.51} & 12.72 & 2.78 & 0.11 \\
\hline
\end{tabular}
\end{table}

Among the evaluated backbones, X3D achieved the highest localization performance, with CorLoc scores of 21.03\%, 9.32\%, and 1.63\% at IoU thresholds of 0.3, 0.5, and 0.7, respectively. Although R(2+1)D-18 achieved the highest Recall, X3D was selected as the backbone of WSPolypNet because the primary objective of this study is spatial polyp localization.

After selecting X3D as the backbone, CAM-based localization was combined with MedSAM2 refinement to construct the complete WSPolypNet framework. Table~\ref{tab:framework_comparison} compares the single-view configuration with the proposed framework.

\begin{table}[!t]
\centering
\caption{Localization performance of the proposed framework.}
\label{tab:framework_comparison}
\begin{tabular}{lcccc}
\hline
Method & Recall & CorLoc@0.3 & CorLoc@0.5 & CorLoc@0.7 \\
\hline
Single-view & 74.18 & 36.87 & 33.72 & 27.94 \\
WSPolypNet  & \textbf{94.51} & \textbf{47.80} & \textbf{43.68} & \textbf{35.01} \\
\hline
\end{tabular}
\end{table}

WSPolypNet achieved CorLoc scores of 47.80\%, 43.68\%, and 35.01\% at IoU thresholds of 0.3, 0.5, and 0.7, respectively, compared with 36.87\%, 33.72\%, and 27.94\% for the single-view configuration. WSPolypNet also achieved a Recall of 94.51\%, indicating improved detection sensitivity compared with the single-view configuration.

To further analyze the effect of multi-view localization with respect to polyp size, we compared the single-view and multi-view configurations separately for large and small polyps. Table~\ref{tab:multiview_comparison} summarizes the results.

\begin{table}[!t]
\centering
\caption{Comparison of single-view and multi-view localization performance for large and small polyps.}
\label{tab:multiview_comparison}
\begin{tabular}{llccc}
\hline
Polyp Size & Setting & CorLoc@0.3 & CorLoc@0.5 & CorLoc@0.7 \\
\hline
\multirow{2}{*}{Large}
& Single-view & \textbf{74.67\%} & \textbf{71.84\%} & \textbf{65.60\%} \\
& Multi-view  & 67.20\% & 65.75\% & 62.34\% \\
\hline
\multirow{2}{*}{Small}
& Single-view & 19.31\% & 16.01\% & 10.45\% \\
& Multi-view  & \textbf{35.39\%} & \textbf{30.97\%} & \textbf{20.29\%} \\
\hline
\end{tabular}
\end{table}

The multi-view strategy substantially improved localization performance for small polyps. In particular, CorLoc at an IoU threshold of 0.5 increased from 16.01\% to 30.97\%. In contrast, for large polyps, CorLoc at the same threshold decreased from 71.84\% to 65.75\%. These results indicate that the proposed multi-view strategy is particularly effective for improving small-polyp localization, although it introduces a trade-off in the localization performance of large polyps.

\section{CONCLUSION}

We proposed WSPolypNet, a weakly supervised framework that combines X3D-based CAM localization with MedSAM2 refinement for polyp localization in colonoscopy videos. WSPolypNet achieved a CorLoc of 43.68\% at an IoU threshold of 0.5, while multi-view localization improved small-polyp CorLoc from 16.01\% to 30.97\%. A current limitation is that the framework selects and propagates a single high-confidence point prompt through MedSAM2, which restricts localization to one polyp even when multiple polyps are present in the same video. Although the current framework is more suitable for annotation assistance than complete annotation replacement, future work will focus on improving localization accuracy, supporting multiple simultaneous polyps, and extending the method to longer colonoscopy videos in which polyps appear only briefly.

\section*{Acknowledgment}
This work was supported by the IITP AI Semiconductor Program and the ANCHOR Program, funded by the Korean government and the Seoul Metropolitan Government (IITP-2023-RS-2023-00256081; 2026-ANCHOR-01-110).

\bibliographystyle{IEEEtran}
\bibliography{references}

\end{document}